# Wilcoxon Nonparametric CFAR Scheme for Ship Detection in SAR Image

Xiangwei Meng

https://orcid.org/0000-0003-3922-977X

*Abstract*—The parametric constant false alarm rate (CFAR) detection algorithms, which are based on various statistical distributions, such as Gaussian, Gamma, Weibull, log-normal, $G^0$, alpha-stable distribution, etc, are most widely used to detect ship targets in SAR images at present. However, the clutter background in SAR images is complicated and variable. When the actual clutter background deviates from the assumed statistical distribution, the performance of the parametric CFAR detector deteriorates; whereas the advantage of the nonparametric detector that its false alarm rate is independent of the background distribution is exhibited. In this work, the Wilcoxon nonparametric CFAR scheme for ship detection in SAR images is proposed and analyzed, and a closed form of the false alarm rate for the Wilcoxon detector to determine the decision threshold is presented. By comparison with several typical parametric CFAR schemes on Radarsat-2, ICEYE-X6 and Gaofen-3 SAR images, the robustness of the ability of the Wilcoxon detector to control the actual false alarm rate at a suitably low level in different detection backgrounds is revealed, and its detection performance for weak ships in rough sea backgrounds is evidently improved. Moreover, the detection speed of the Wilcoxon detector is fast, and it has simple hardware implementation.

## I. INTRODUCTION

THE synthetic aperture radar (SAR) has been widely used for earth remote sensing for many years. It provides high resolution images independent of daylight, cloud coverage and weather conditions. Ship detection in SAR images is an important application of SAR remote sensing and has attracted increased interest from experts in this field. It can be used for ship traffic monitoring, port security, antipiracy and military applications. Among the many methods available for detecting ship targets in SAR images, the constant false alarm rate (CFAR) scheme is the most extensively and effectively applied technique. The famous detector is the two-parameter CFAR detector [1], which uses the estimated mean and standard deviation of the clutter samples within a square band window centred around

This work was supported by the National Natural Science Foundation of China under Grant 62171402. *(Corresponding author: Xiangwei Meng)*.

Xiangwei Meng is with the Department of Electrical and Electronic Engineering, Yantai Nanshan University, Yantai 265713, China (e-mail: mengxw163@ sohu.com).

the pixel under test to set an adaptive threshold. The two-parameter CFAR detector can maintain a constant false alarm rate under the assumption of a Gaussian background.

When the radar resolution increases, the distribution of the SAR clutter deviates from the Gaussian distribution and shows a long tail of the distribution. To improve the ship detection performance in SAR images, various statistical distributions have been used to model the SAR clutter backgrounds [2-11], such as K, the alpha-stable distribution, $G^0$, log-normal, Gamma, Rayleigh, Rayleigh mixtures distribution, etc. In [2], the simulated annealing method was used to segment the high-resolution SAR images into different regions with homogeneous characteristics, and the CA-CFAR based on the K distribution was performed inside each homogeneous region. In [3], the *K*-means program was used to segment an SAR image into *N* sub-images, and the CFAR detection based on the alpha-stable distribution was carried out. In [4], an adaptive and fast CFAR algorithm based on the $G^0$ distribution for ship detection in SAR images was proposed. In [5], the log-normal distribution was used to model the clutter background in SAR images, and a multilayer CFAR detection for ship target in SAR images was proposed, which repeatedly detects and eliminates the target pixels. To reduce the influence of multiple targets on the estimation of background statistics, the truncated statistics (TS) CFAR method [6] based on the Gamma distribution for ship detection in single-look intensity and multi-look intensity SAR data was proposed. In [7], a novel and fast target detection approach called RmSAT-CFAR for SAR images was presented; this approach uses Rayleigh mixtures for clutter modeling. An outlier-robust CFAR detector that aims at elevating the detection performance in multiple target situations was proposed in [8], which assumes a Gaussian distribution for the sea clutter in SAR images. In [9], a robust constant false alarm rate detector based on censored harmonic averaging (CHA-CFARD) was proposed for dense interfering target situations, and its effectiveness was verified via ship detection in TerraSAR-X images. In [10], a robust CFAR detector based on bilateral-trimmed-statistics called the BTS-RCFAR for ship detection in SAR images was proposed; this detector uses the log-normal distribution to fit the sea clutter in SAR images. In [11], an automatic identification system (AIS) data-aided Rayleigh constant false alarm rate (AIS-RCFAR) ship detection algorithm was proposed. The Rayleigh distribution is used to model clutter in SAR images, and a clutter trimming method is designed with an adaptive trimming depth aided by AIS data to effectively eliminate the high-intensity outliers in the local background window.

In modern radar systems, the use of constant false alarm rate techniques is necessary to keep false alarms at a suitably low rate in an a priori unknown time-varying and spatially nonhomogeneous environment [12], [13]. Moreover, the design false alarm rate of a CFAR scheme and its actual false alarm rate occurring in the detection results may be vastly different. Therefore, to make a fair and accurate performance comparison between the different CFAR schemes for ship detection in SAR images, first, their ability to control the actual false alarm rates in different detection backgrounds must be evaluated. The second step is to analyze the detection performance of the CFAR algorithms under the same or similar actual false alarm rates. Nevertheless, the performances of the parametric CFAR detectors with different assumed distributions are usually compared under the same design false alarm rate, which is unfair. Indeed, a performance comparison of the parametric CFAR detectors under the same design false alarm rate is feasible for the same statistical distribution, but the performances of the CFAR schemes with different assumed distributions should be compared under the same or similar actual false alarm rates.

In fact, the distribution of the SAR image background depends on many factors, such as the radar frequency, resolution, polarization, incidence angle, wind speed, sea state, etc. Even if the distribution of the clutter data is known at some time, uncontrollable phenomena may cause changes such that at a later time, the clutter distribution is vastly different. The clutter background in SAR images is variable and complicated. When the actual clutter background deviates from the assumed statistical distribution, the performance of the parametric CFAR detectors deteriorates, whereas the nonparametric CFAR procedure exhibits the advantage that it can maintain a constant false alarm rate despite changes in the underlying data distribution.

The primary motivation of this work is to propose a new path for performing ship target detection in SAR images, that is, the application of the well-known Wilcoxon nonparametric detector to ship target detection in SAR images. The effectiveness of the ability of the Wilcoxon nonparametric detector to control actual false alarms against different backgrounds and its ability to detect ships in SAR images will be validated in this work. In [14], the Wilcoxon two-sample nonparametric detector, which is usually termed the rank sum (RS) nonparametric detector in the literature, was applied to low-resolution radar target detection. The generalized sign test detector [15] was proposed at the same time, and it has the same form as the RS nonparametric detector. A closed-form expression of the false alarm rate of the RS nonparametric detector in a homogeneous background was derived by Akimov (in Russian) and was reported by M. Sekine and Y. H. Mao [16]. The analytical expressions of the false alarm rate for the RS and rank quantization (RQ) nonparametric detector [17] in a nonhomogeneous background were derived in [18] and [19], respectively.

The Wilcoxon nonparametric CFAR scheme for ship detection in SAR images is analyzed in this work. A description of the Wilcoxon nonparametric CFAR algorithm for ship detection in SAR images is given in Section II. The ability of the Wilcoxon nonparametric CFAR method to detect ships in Radarsat-2, ICEYE-X6 and Gaofen-3 SAR images is analyzed in Section III. We discuss the ability of the Wilcoxon nonparametric CFAR to control the actual false alarm rates in these 3 SAR images, and compare its detection probability to that of the two-parameter CFAR, the Weibull-CFAR, the TS-CFAR and the AIS-RCFAR under the same or similar false alarm rates in Section IV. Finally, we summarize and discuss the results obtained.

## II. DESCRIPTION OF THE WILCOXON NONPARAMETRIC CFAR DETECTOR

In a modern radar signal processing system, the target CFAR detection is commonly performed using the sliding window technique [20], [21], [22]. The data available in the reference window enter into an algorithm for the calculation of the decision threshold. For ship detection in SAR images with the Wilcoxon nonparametric CFAR detector, a sketch of the two-dimensional sliding window is shown in Fig. 1. As shown in the figure, the cells (pixels) under test are at the center of a defined local region, the cells in the boundary stencil are the reference cells, and the cells between the test cells and the reference cells are the guard cells. The guard area ensures that no target signal energy spills over into the reference cells. Let $Y_1, \cdots, Y_n$ denote the reference cells in background region, whereas $X_1, \cdots, X_m$ represent the test cells used to decide whether a target is present.

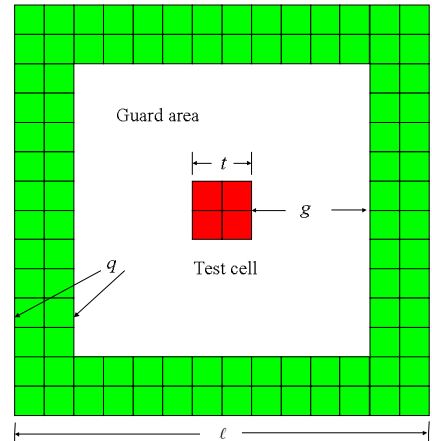

**Fig. 1.** The sliding window of the Wilcoxon nonparametric CFAR detector.

We assume that $X_1, \cdots, X_m$ and $Y_1, \cdots, Y_n$ are independent and identically distributed (IID) random samples with probability density functions (PDFs) $f(x)$ and $g(x)$, respectively. Under the null hypothesis $H_0$, the test samples $X_1, \cdots, X_m$ and the reference samples $Y_1, \cdots, Y_n$ are IID and have the same distribution. We also assume that two samples are drawn from continuous distributions, so that the possibility $X_i = Y_j$ for some $i$ and $j$ need not to be considered. Setting $X_{m+j} = Y_j$, $j = 1, \cdots, n$, and $N = m + n$, we obtain a combined sample set $X_1, \cdots, X_N$. The combined samples $X_1, \cdots, X_N$ are first rank-ordered according to increasing magnitude. The sequence thus obtained is

$$X_{(1)} \leq X_{(2)} \leq \cdots \leq X_{(N)} \qquad (1)$$

The sequence given in (1) is called the order statistic. The indices in parentheses indicate the rank-order number. Let $R_i \ (i = 1, \cdots, N)$ denote the rank of the observation $X_i$ in the ordered sequence (1). We observe that $R_1, \cdots, R_m$ are the ranks corresponding to the test samples $X_1, \cdots, X_m$ in the combined samples $X_1, \cdots, X_N$.

To decide whether a radar target exists in the test cells, the Wilcoxon nonparametric CFAR detector is applied to this situation. The Wilcoxon test statistic is in the form of

$$S_{m,n} = \sum_{i=1}^{m} R_i \qquad (2)$$

If $S_{m,n}$ exceeds a decision threshold $T_W$ ($S_{m,n} \geq T_W$), a target is declared to be present in the test cells. On the other hand, if it is less than $T_W$ ($S_{m,n} < T_W$), the target is absent. The decision threshold $T_W$ is chosen to yield the desired or design false alarm probability $P_{FA}$.

**Theorem 1** [23]: *Let $S_{m,n}$ be the two-sample Wilcoxon test statistic, for the samples $X_1, \cdots, X_m$ and $Y_1, \cdots, Y_n$. Under the null hypothesis $H_0$, the probability of the Wilcoxon test statistic taking $S_{m,n} = k$ is*

$$P\{S_{m,n} = k\} = \frac{\pi_{m,n}(k)}{\binom{N}{m}} \qquad (3)$$

*for $k = \frac{1}{2}m(m+1), \frac{1}{2}m(m+1)+1, \cdots, \frac{m(m+2n+1)}{2}$, where $\pi_{m,n}(k)$ satisfy the recurrence formula*

$$\pi_{m,n}(k) = \pi_{m,n-1}(k) + \pi_{m-1,n}(k-m-n) \qquad (4)$$

*with the initial and boundary conditions:*

$$\pi_{i,0}\left(\frac{1}{2}i(i+1)\right) = 1 ; \qquad (5)$$

$$\pi_{i,0}(k) = 0, \text{ for } k \neq \frac{1}{2}i(i+1), \ i = 1, \cdots, m ; \qquad (6)$$

$$\pi_{0,j}(0) = 1 ; \qquad (7)$$

$$\pi_{0,j}(k) = 0, \text{ for } k \neq 0, \ j = 1, \cdots, n ; \qquad (8)$$

*The notation $\pi_{m,n}(k)$ in (3) is the number of distinguishable arrangements of samples $X_1, \cdots, X_m$ and $Y_1, \cdots, Y_n$, regardless of their indices, in $X_{(1)}, \cdots, X_{(N)}$, such that $S_{m,n} = \sum_{i=1}^{m} R_i = k$.*

For readability, the proof of Theorem 1 in [23] is also cited here.

**Proof:** Let us denote by $\pi_{m,n}(k)$ the number of arrangements of observations $\{X_i\}$ and $\{Y_j\}$, regardless of their indices, in $X_{(1)}, \cdots, X_{(m+n)}$, such that $S_{m,n} = \sum_{i=1}^{m} R_i = k$. Now, (3) is obvious, and (4) follows easily by discarding the observations $X_{(m+n)}$. Actually, if $X_{(m+n)}$ is some $Y_j$, we get $S_{m,n-1} = k$; and if $X_{(m+n)}$ is some $X_i$, we get $S_{m-1,n} = k - m - n$.

Under the null hypothesis $H_0$, if the Wilcoxon test statistic $S_{m,n}$ is greater than a decision threshold $T_W$, a false alarm occurs. Therefore, the false alarm probability of the Wilcoxon nonparametric CFAR scheme for ship detection in SAR images is expressed as

$$P_{FA} = \sum_{k=T_W}^{m(m+2n+1)/2} P\{S_{m,n} = k\} = \sum_{k=T_W}^{m(m+2n+1)/2} \frac{\pi_{m,n}(k)}{\binom{N}{m}} \qquad (9)$$

The derivation of the design false alarm rate $P_{FA}$ of the Wilcoxon nonparametric CFAR method for ship detection in SAR images is irrelevant to the distribution of background clutter. Therefore, the Wilcoxon nonparametric CFAR scheme for ship detection in SAR images can maintain a constant false alarm rate regardless of the distribution of the clutter background. This is what we call the distribution-free property of a nonparametric CFAR detector.

Since the Mann-Whitney nonparametric detector is equivalent to the Wilcoxon scheme [24] and is easy to implement, we replace the Wilcoxon scheme with the Mann-Whitney detector in our analysis. The test statistic of the Mann-Whitney nonparametric detector is defined as

$$R_{MW} = \sum_{i=1}^{m} \sum_{j=1}^{n} u(x_i - y_j) \qquad (10)$$

where the $u(t)$ is the unit step function and $u(t) = \begin{cases} 1 & t \geq 0 \\ 0 & t < 0 \end{cases}$.

If the test statistic $R_{WM}$ of the Mann-Whitney detector is greater than the decision threshold $T_{MW}$, a declaration about the presence of a target in the test window is made; otherwise, no target exists. The relationship between the test statistic of the Wilcoxon detector and that of the Mann-Whitney detector is $S_{m,n} = R_{MW} + \frac{1}{2}m(m+1)$. The decision threshold $T_{MW}$ of the Mann-Whitney nonparametric detector for a design false alarm rate $P_{FA}$ can also be obtained by substituting $T_{MW} = T_W - \frac{1}{2}m(m+1)$. Actually, the design false alarm rate determined by an analytical expression such as (9) of a CFAR scheme may be vastly different from the actual false alarm rate in the detection results. For ease of distinction, the design false alarm rate determined by the analytical expression is denoted as $P_{FA}$, and the actual false alarm rate occurring in the detection results is represented by $P_{fa}$ in the following.

## III. Experimental Results and Performance Analysis

The performance of the Wilcoxon nonparametric CFAR scheme for ship detection in SAR images is verified by 3 SAR image scenes: SAR ship detection in a calm sea background, SAR ship detection in a rough sea background and weak ship detection in a rough SAR background. The data were collected by Radarsat-2, ICEYE-X6 and Gaofen-3, respectively. The detection results of the Wilcoxon nonparametric CFAR detector for ships in 3 SAR scenes are presented and analyzed in the following Sections.

### A. SAR ship detection in a calm sea background

#### 1) SAR Image #1 collected by Radarsat-2

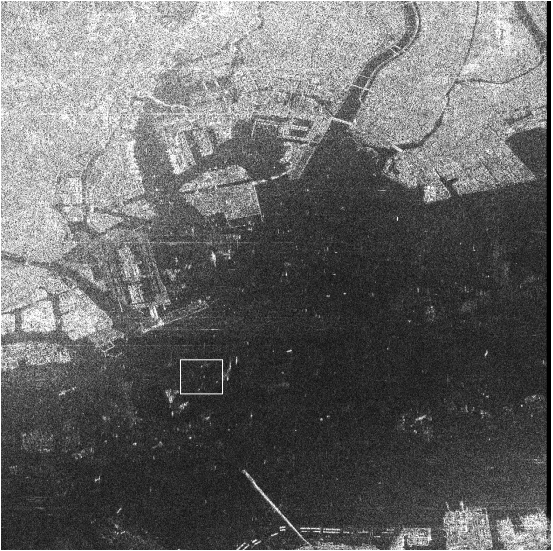

**Fig. 2.** The SAR image was acquired by the C-band Radarsat-2 on March 8, 2010, around a seaport near Tokyo, Japan.

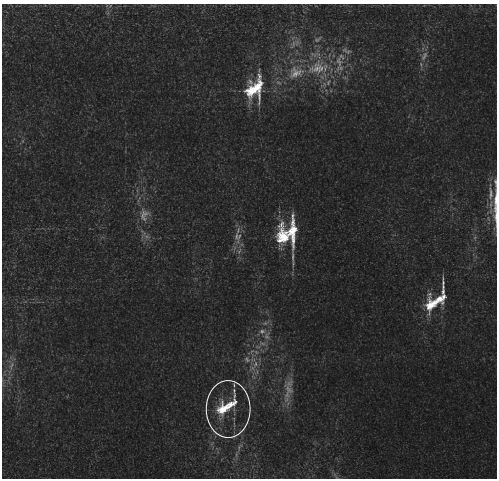

**Fig.3.** SAR image #1 from Radarsat-2.

The first SAR image data examined here were acquired by the C-band Radarsat-2 in the standard mode on March 8, 2010, around a seaport near Tokyo, Japan. The polarization mode is HH. The acquired SAR image is shown in Fig. 2. The SLC data cover a scene of about 20 km by 20 km, which corresponds to an image of size $16036 \times 11955$. The azimuth and range resolutions are 3 m, respectively. To validate the detection performance of the Wilcoxon nonparametric detector, a slice marked by a white rectangle in Fig. 2 is cut, which is SAR image #1 shown in Fig. 3. SAR image #1 is $1001 \times 901$ in size, 4 ships are visible, and the extended sidelobes caused by the strong reflections of the ship body can be clearly observed. The sea surface in the SAR image is calm, but some speckles exist in the background.

Although the Wilcoxon nonparametric detector can detect targets in a nonGaussian background at a constant false alarm rate without assuming a known distribution, to compare the results to those of other parametric CFAR schemes, a statistical distribution analysis of the SAR image data is necessary. In the design stage of a CFAR detection scheme, a relatively low design value of the false alarm rate $P_{FA}$ for conventional radar is usually determined in a homogenous background. Therefore, the pixels from 4 targets in SAR image #1 should be discarded first, in an attempt to accurately model the clutter background. The statistical analysis of SAR image #1 is shown in Fig. 4. A histogram of the pixels of the clutter background in SAR image #1 is depicted, and the PDF curves of the Gaussian, Weibull, K, Gamma and Rayleigh distributions used to fit the background clutter data are also included. Two cases of the Gamma distribution are presented in Fig. 4; one is that the shape parameter of the Gamma distribution is estimated by the maximum likelihood method, which is denoted by "Gamma", and the other is that the shape parameter of the Gamma distribution is replaced by the equivalent number of looks (ENL), which is represented by "Gamma-ENL". It can be seen that the main body of the PDF of the K distribution, that of the Weibull distribution and that of the Rayleigh distribution fit the histogram of SAR image #1 relatively well, and the worst case is obtained by the Gaussian distribution. The main bodies of the PDFs of both Gamma distributions deviate from the histogram of SAR image #1 to a certain extent.

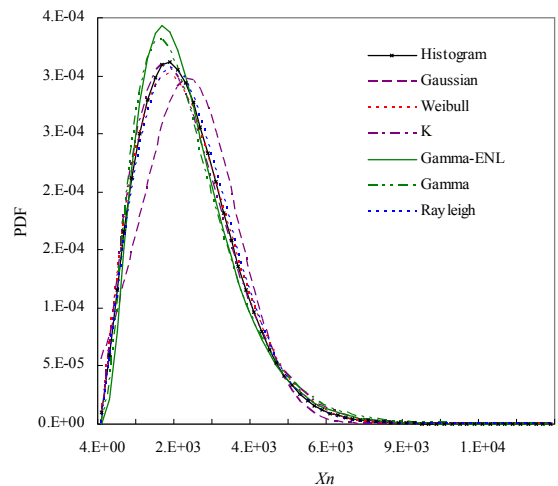

**Fig. 4.** The analysis of the statistical distributions on SAR image #1.

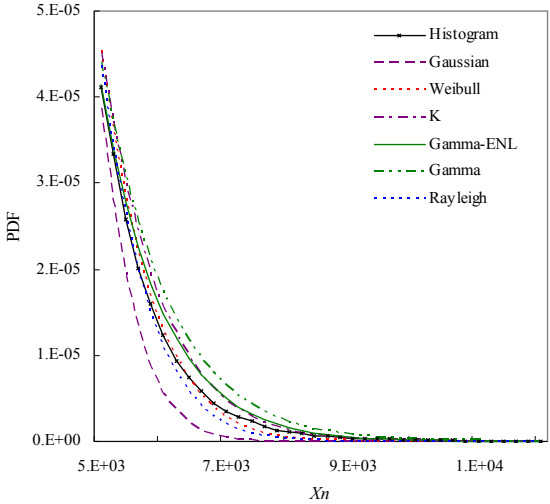

**Fig. 5.** The tails of the statistical distributions of the SAR image #1.

In fact, the tail of the PDF on the right side determines the false alarm performance of a CFAR detection scheme. To inspect the difference between the tails of the PDFs of these distributions in detail, the tails of these distributions are illustrated in Fig. 5 again. The tail of the Gaussian distribution significantly deviates from that of the histogram of SAR image #1. As the gray value $X_n$ becomes large, the tail of the Weibull distribution and that of the Rayleigh distribution cross the histogram of SAR image #1 and then are lower than it. The tail of the K distribution and that of the Gamma distribution with ENL almost come together and are slightly higher than that of the histogram, whereas the tail of the Gamma distribution with maximum likelihood estimation is even higher than them. The shape parameter of the Gamma distribution with maximum likelihood estimation is 2.99, and the ENL to replace the actual shape parameter is 3.33. It can be seen that although these PDF curves of several statistical distributions approach the histogram of SAR image #1 to varying degrees, all of them cannot coincide with it completely.

### 2) Performance analysis of the Wilcoxon nonparametric detector

The Wilcoxon detector takes a sliding window of size $\ell \times \ell = 68 \times 68$, a width $g$=30 of the guard area and a test window of size $t \times t$=2×2. The $\ell$, $g$ and $t$ are shown in Fig. 1, respectively. To avoid the influence of the pixels under test on the background clutter estimation, the width of the guard area is longer than the length of the largest ship in the SAR image. For the Wilcoxon nonparametric detector in our experiments, the boundary clutter samples of $q$=3 rows/columns in the sliding window form the reference samples to set the detection threshold, and the total number of reference samples is 780.

To make a comparison with the parametric CFAR schemes, we consider the two-parameter CFAR, the Weibull-CFAR [25], the TS-CFAR [6] and the AIS-RCFAR [11] as a reference. The two-parameter CFAR, the Weibull-CFAR, the TS-CFAR and the AIS-RCFAR operate in a conventional pixel by pixel detection approach. Only a single pixel in the test window is detected at a time for them, whereas the Wilcoxon detector slides 2 pixels at a time. The two-parameter CFAR, the Weibull-CFAR and the TS-CFAR use the same width $g$=30 of the guard area, but the AIS-RCFAR does not use the guard window. Since a total of 248 clutter samples in the reference window are sufficient for the maximum likelihood estimation of distribution parameters, a single boundary row/column $q$=1 of clutter samples in the reference window is adopted for the two-parameter CFAR, the Weibull-CFAR and the TS-CFAR. Here, a sliding window of size $\ell \times \ell = 63 \times 63$ is taken for the two-parameter CFAR, the Weibull CFAR, the TS-CFAR and the AIS-RCFAR. The TS-CFAR scheme takes a truncation ratio [6] at $R_t$ =10% , and it uses the ENL to replace the actual shape parameter. The parameter λ of the AIS-RCFAR is λ=2.0 [11].

The detection results of the two-parameter CFAR are shown in Figs. 6(a), (b) and (c) for the design false alarm rates $P_{FA}$=10⁻⁵, 10⁻⁶ and 10⁻⁷, respectively. The actual false alarm rates $P_{fa}$ for the two-parameter CFAR in the detection results are also included in the subtitles of Fig.6. We utilize the maximum-likelihood estimation to obtain the mean and the variance of the background clutter and then set the adaptive detection threshold. The detection results of the Weibull-CFAR are shown in Figs. 7(a), (b) and (c) for the design false alarm rates $P_{FA}$=10⁻³, 10⁻⁴ and 3×10⁻⁵, respectively. Interestingly, the detection results of the two-parameter CFAR at $P_{FA}$=10⁻⁷ are similar to those of the Weibull CFAR at $P_{FA}$=3×10⁻⁵. This is because the two-parameter CFAR assumes a Gaussian distribution for the background clutter, whereas the actual clutter in the SAR image obeys a nonGaussian distribution with a long tail; thus, a higher detection threshold for the two-parameter CFAR at $P_{FA}$=10⁻⁷ is required to achieve the similar detection effects as those of the Weibull-CFAR at $P_{FA}$=3×10⁻⁵. Therefore, a comparison of detection performance between the CFAR detectors with different assumed distributions under the same design false alarm rate $P_{FA}$ is unfair.

In modern radar systems, the use of constant false alarm rate techniques is necessary to keep false alarms at a suitably low rate in an a priori unknown time-varying and spatially nonhomogeneous environment [12], [13]. Moreover, the design false alarm rate of a CFAR scheme and its actual false alarm rate occurring in detection results may be vastly different. Under the condition of the same statistical distribution, a performance comparison between different CFAR detectors with the same design false alarm rate $P_{FA}$ is feasible, but a performance comparison between the CFAR schemes with different assumed distributions under the same design false alarm rate $P_{FA}$ is unfair. Therefore, in order to make a fair and accurate performance comparison, it is necessary to compare the detection performances of different CFAR schemes under the same or similar actual false alarm rates $P_{fa}$. This approach will be of more practical interest.

The actual false alarm rate $P_{fa}$ is defined as the ratio of the number of false alarms to the total number of pixels in the SAR image background, that is,

$$P_{fa} = \frac{N_{fa}}{N_c} = \frac{N_{fa}}{P_1 \times P_2 - \sum_{i=1}^{Q} N_s^{(i)}} \quad (11)$$

where $N_{fa}$ is the number of false alarms occurring in the detection results and $N_c$ is the total number of pixels in the SAR image background. $P_1$ and $P_2$ are the length and width of the SAR image, respectively. $Q$ is the number of ships in the SAR image. $N_s^{(i)}$ is the number of pixels in the $i$th ship body. We use the ellipse to fit the ship body, and the major and minor axes ($a$ and $b$) of the ellipse are estimated; then, $N_s^{(i)}$ is determined by the area of the ellipse $\pi ab / 4$. For SAR image #1 shown in Fig. 3, the length and width of the image slice are 1001 and 901, respectively. The total number of pixels from the 4 ship bodies is 1583; thus, $N_c$=1001×901-1583=900318. The design false alarm rate $P_{FA}$ and the actual false alarm rate $P_{fa}$ for the two-parameter CFAR are given in the subtitles of Fig. 6, as are the other CFAR schemes. Although the design false alarm rates of the two-parameter CFAR and the Weibull-CFAR in Fig. 6(c) and Fig. 7(c) are $P_{FA}$=10$^{-7}$ and $P_{FA}$=3×10$^{-5}$, respectively, they exhibit similar actual false alarm rates of $P_{fa}$=1.13×10$^{-4}$ and $P_{fa}$=1.18×10$^{-4}$.

The detection results of the TS-CFAR scheme with the truncation ratio $R_t = 10\%$ for SAR image #1 at the design false alarm rates $P_{FA}$=10$^{-2}$, 10$^{-3}$ and 10$^{-4}$ are shown in Figs. 8(a), (b) and (c), respectively. Since the TS-CFAR assumes a Gamma distribution for the clutter background and the tail of the Gamma distribution is slightly greater than that of the Weibull distribution, the TS-CFAR at $P_{FA}$=10$^{-4}$ has similar detection effects to those of the Weibull-CFAR at $P_{FA}$=3×10$^{-5}$. However, most false alarms of the TS-CFAR are caused by the sidelobes of the ship. This is different from the case of the two-parameter CFAR and the Weibull-CFAR.

The detection results of the AIS-RCFAR for SAR image #1 at the design false alarm rates $P_{FA}$=10$^{-4}$, 10$^{-5}$ and 3×10$^{-6}$ are shown in Figs. 9(a), (b) and (c), respectively. The detection results of the Wilcoxon nonparametric detector for SAR image #1 at the design false alarm rates $P_{FA}$=10$^{-6}$, 10$^{-7}$ and 10$^{-8}$ are shown in Figs. 10(a), (b) and (c), respectively. As shown in Fig. 6(c) through Fig. 10(c), although the design false alarm rates $P_{FA}$ of the two-parameter CFAR, the Weibull-CFAR, the TS-CFAR, the AIS-RCFAR and the Wilcoxon nonparametric detector are different, they obtain similar actual false alarm rates $P_{fa}$ near an acceptable level of 10$^{-4}$. For comparison, the values of $N_{fa}$, $N_c$, $P_{fa}$ and $P_{FA}$ for the detection results in Fig. 6(c) through Fig. 10(c) are included in Table I.

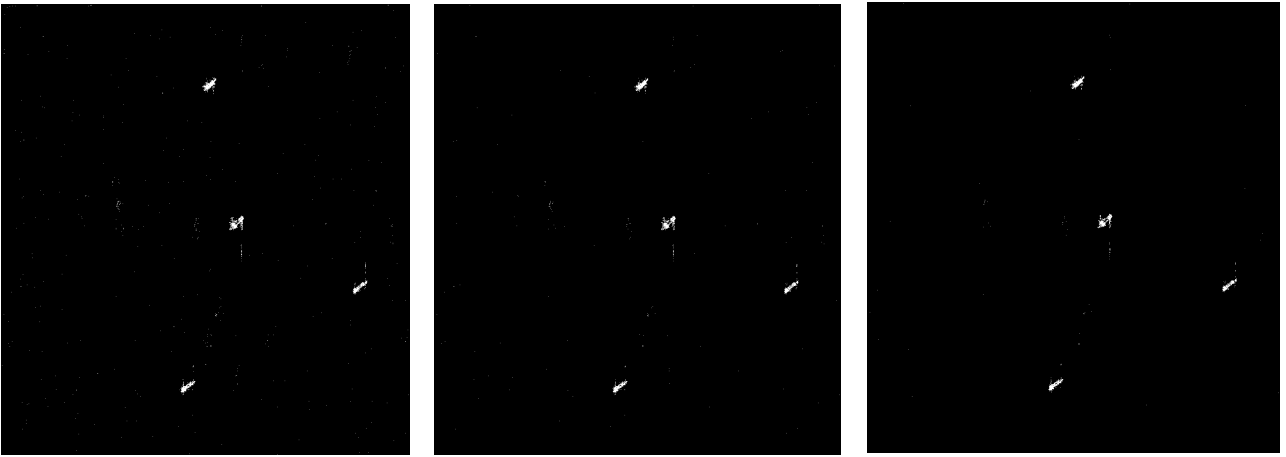

(a) $P_{FA}$=10$^{-5}$, $P_{fa}$=4.93×10$^{-4}$      (b) $P_{FA}$=10$^{-6}$, $P_{fa}$=2.17×10$^{-4}$      (c) $P_{FA}$=10$^{-7}$, $P_{fa}$=1.13×10$^{-4}$

**Fig. 6.** The detection results of the two-parameter CFAR on SAR image #1.

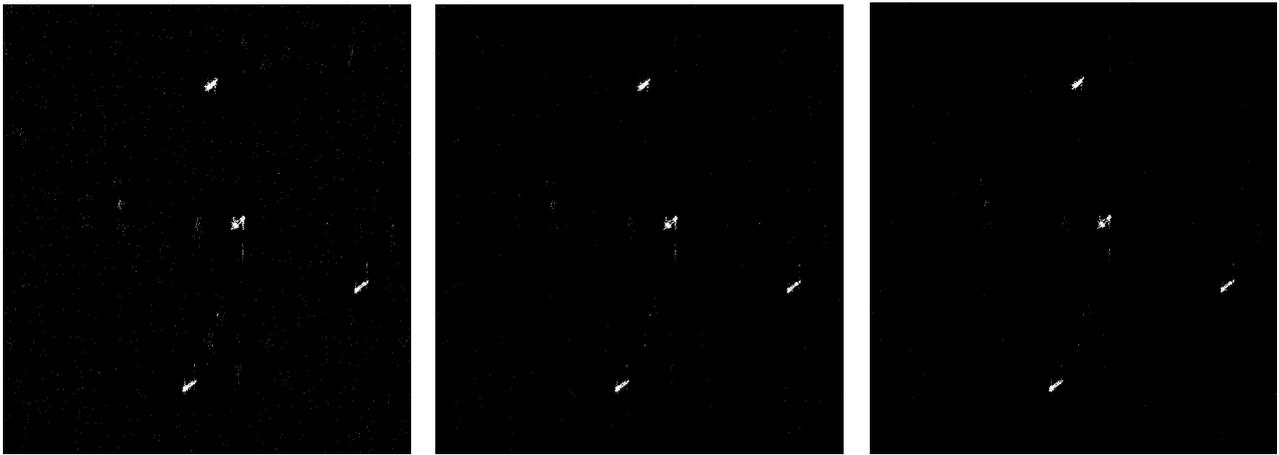

(a) $P_{FA}=10^{-3}$, $P_{fa}=1.3\times10^{-3}$      (b) $P_{FA}=10^{-4}$, $P_{fa}=2.4\times10^{-4}$      (c) $P_{FA}=3\times10^{-5}$, $P_{fa}=1.18\times10^{-4}$

**Fig. 7.** The detection results of the Weibull-CFAR on SAR image #1.

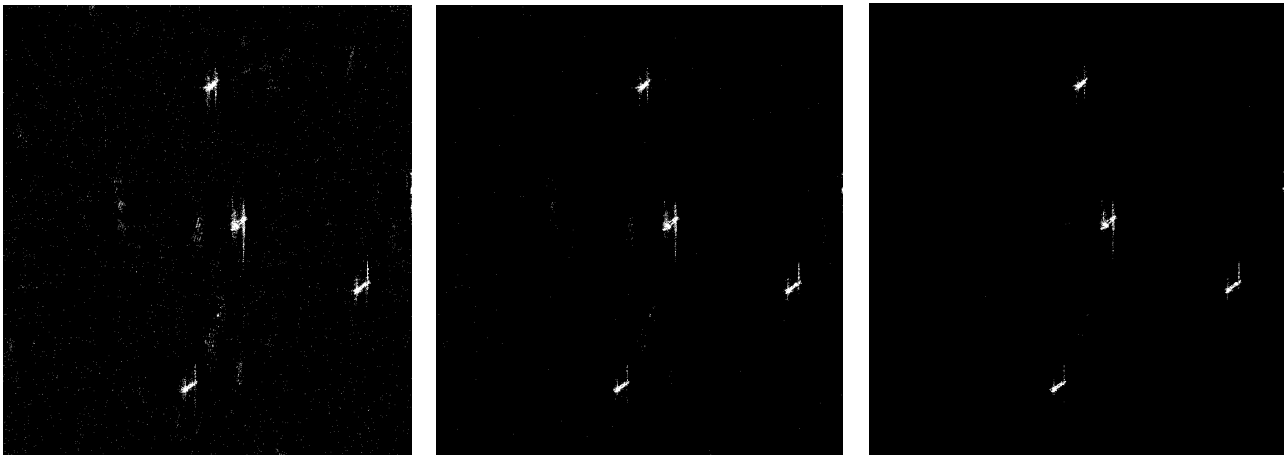

(a) $P_{FA}=10^{-2}$, $P_{fa}=3.3\times10^{-3}$      (b) $P_{FA}=10^{-3}$, $P_{fa}=3.09\times10^{-4}$      (c) $P_{FA}=10^{-4}$, $P_{fa}=1.36\times10^{-4}$

**Fig. 8.** The detection results of the TS-CFAR scheme on SAR image #1.

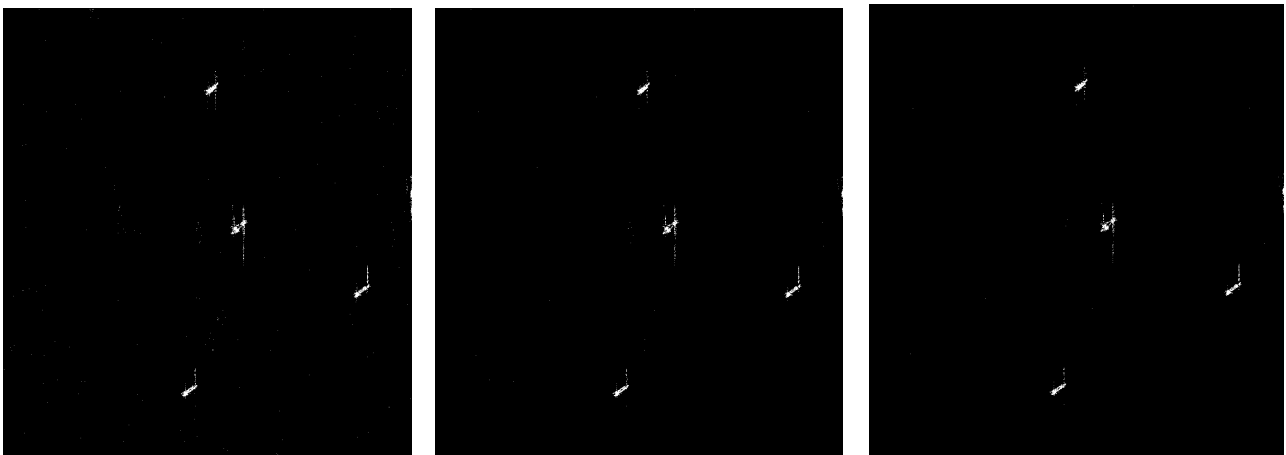

(a) $P_{FA}=10^{-4}$, $P_{fa}=3.32\times10^{-4}$      (b) $P_{FA}=10^{-5}$, $P_{fa}=1.56\times10^{-4}$      (c) $P_{FA}=3\times10^{-6}$, $P_{fa}=1.31\times10^{-4}$

**Fig. 9.** The detection results of the AIS-RCFAR scheme on SAR image #1.

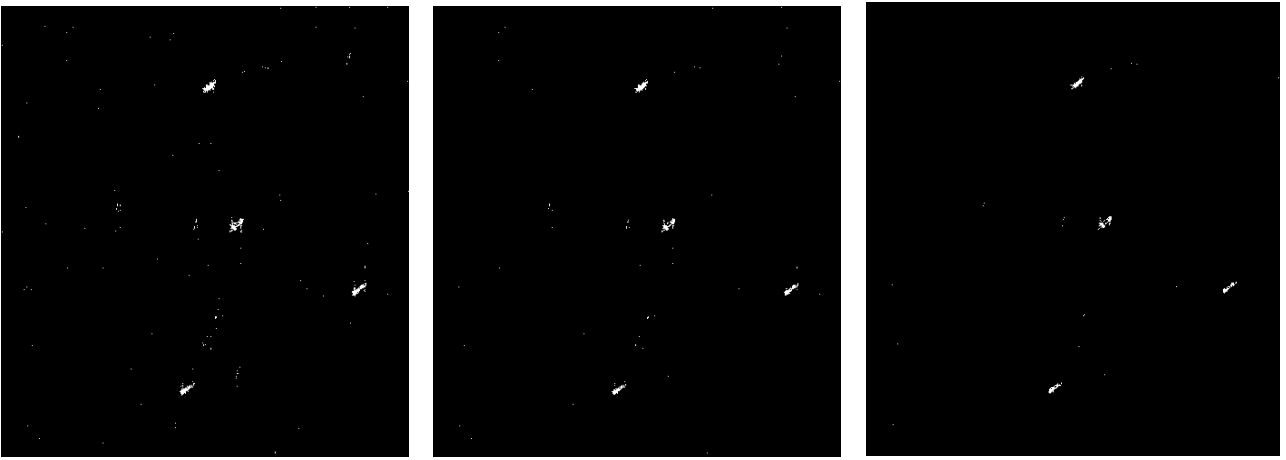

(a) $P_{FA}=10^{-6}$, $P_{fa}=1.29\times10^{-4}$      (b) $P_{FA}=10^{-7}$, $P_{fa}=0.6\times10^{-4}$      (c) $P_{FA}=10^{-8}$, $P_{fa}=0.29\times10^{-4}$

**Fig. 10.** The detection results of the Wilcoxon nonparametric detector on SAR image #1.

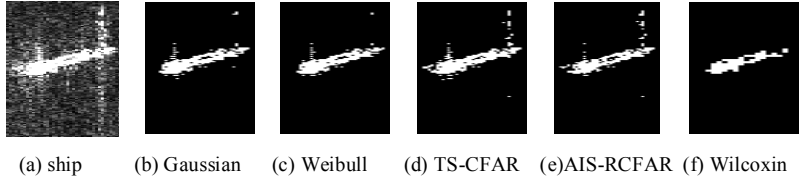

(a) ship    (b) Gaussian    (c) Weibull    (d) TS-CFAR    (e)AIS-RCFAR (f) Wilcoxin

**Fig. 11.** The detection effects of several detectors on the ship of SAR image #1.

In Fig. 6(c) through Fig. 10(c), the detection effects of the two-parameter CFAR, the Weibull CFAR, the TS-CFAR, the AIS-RCFAR and the Wilcoxon nonparametric detector for the ship in the white circle in SAR image #1 are displayed in Fig. 11 again. For strong ship targets in the SAR images, all of these detectors can detect most of the pixels in the ship body under a similar actual false alarm rate of $10^{-4}$ level, and the Wilcoxon nonparametric detector can suppress the false alarms resulting from the sidelobe of the ship body to some extent. This advantage of the Wilcoxon nonparametric detector lies in the fact that the Wilcoxon nonparametric detector takes a $t\times t=2\times 2$ test window such that the wirelike sidelobes can be suppressed to some extent.

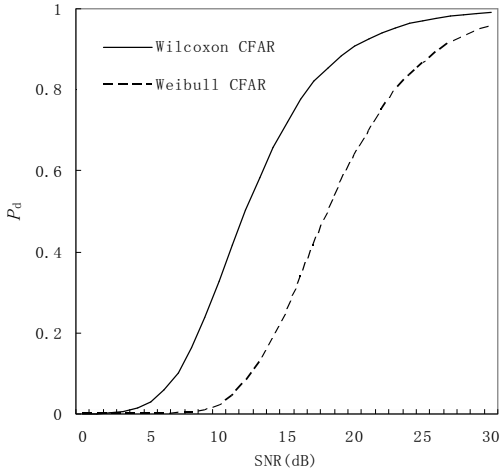

**Fig. 12.** The detection performance of the Wilcoxon nonparametric detector.

Naturally, a question arises now. The false alarm rate of the Wilcoxon nonparametric detector is independent of the background clutter distribution. Why does the Wilcoxon nonparametric detector use a lower design false alarm rate $P_{FA}=10^{-8}$ to reach the actual false alarm rate of $10^{-4}$ level? The cause behind this phenomenon is that the Wilcoxon nonparametric detector takes 4 pixels in the test window to perform target detection in the SAR image, and the speckles in the SAR image appear to be cotton-like and slice-like. It is more likely that these speckles will be detected as false alarms by the Wilcoxon nonparametric detector.

In Fig. 12, a comparison of the detection performance of the Wilcoxon nonparametric detector with 4 test pixels and 780 reference samples to that of the Weibull-CFAR with a single pixel under test and 248 reference samples is made by Monte Carlo simulation under the same design false alarm rate $P_{FA}=10^{-5}$. We assume that the background clutter follows the Weibull distribution and that the intensity of the pixels of the ship target obeys the Swerling II fluctuation. In the simulation, the target plus the background clutter in the test cells are added as vectors. The Wilcoxon nonparametric detector yields better detection performance than does the Weibull-CFAR for the same design false alarm rate $P_{FA}=10^{-5}$. This means that while the pixels of the ship target are detected by the Wilcoxon nonparametric CFAR detector at a high detection probability, the speckles in the SAR image are also easily detected. Therefore, a higher detection threshold corresponding to a lower design false alarm rate $P_{FA}=10^{-8}$ is required by the Wilcoxon nonparametric detector to suppress the resulting false alarms in a specific SAR image scene.

### 3) The time cost of the Wilcoxon nonparametric detector

The time cost is an important aspect of the performance of the CFAR detection scheme. It determines whether an algorithm can be successfully applied in practical application. The CFAR detection schemes examined in this work are implemented in MATLAB, and run on a PC with an Intel Core i7 CPU @ 3.0 GHz and 16 GB of memory. The execution times $T_s$ of the two-parameter CFAR (denoted as TP-CFAR in Table I), the Weibull-CFAR, the TS-CFAR, the AIS-RCFAR and the Wilcoxon nonparametric detector on SAR image #1 for Fig. 6(c) through Fig. 10(c) are also included in Table I. The time cost of the Wilcoxon nonparametric detector on SAR image #1 is the least among these CFAR schemes. It reduces more than half of that of the two-parameter CFAR. The time cost of the TS-CFAR is the longest. This is because the Wilcoxon nonparametric detector does not assume a known form for background distribution; thus, it avoids the large and complicated computations needed for the parameter estimation of the parametric CFAR detectors.

### B. SAR ship detection in a rough sea background

### 1) SAR Image #2 collected by ICEYE-X6

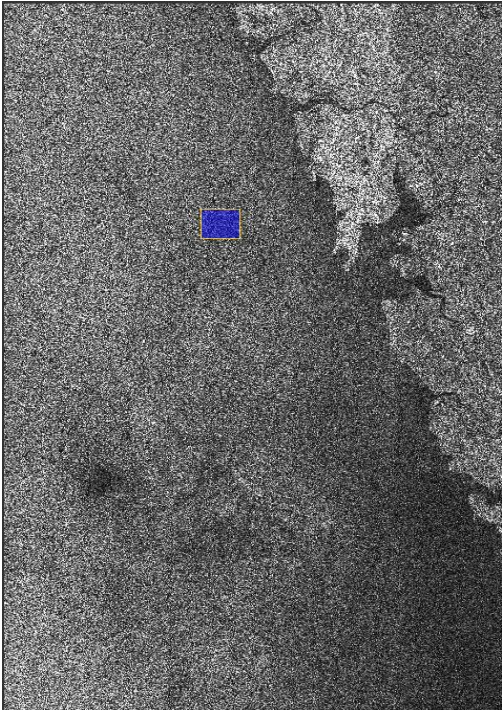

**Fig. 13.** The SAR image was acquired by the ICEYE-X6 satellite on 24 June 2022 at night, covering an area of 20,000 sqkm along the coastlines of Maharashtra, India.

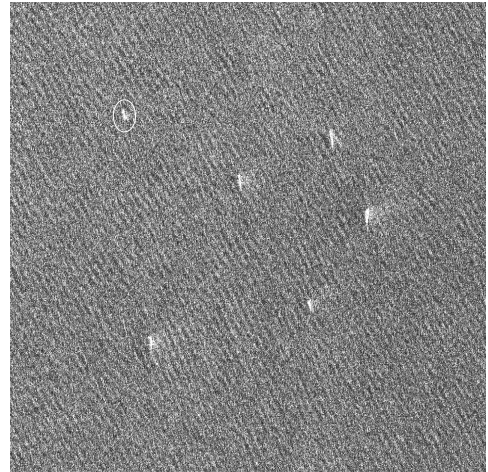

**Fig. 14.** SAR image #2 from ICEYE.

The SAR image [26] shown in Fig. 13 was acquired by the ICEYE-X6 satellite on 24 June 2022 at night, which operates in the X-band, in the Scan mode and the VV polarization, covering an area of 20,000 sqkm along the coastline of Maharashtra, India. The data were Ground Range Detected (GRD) in GeoTiff format, and the image had a resolution better than 15 m. SAR image #2 shown in Fig. 14 is a slice (marked with a blue rectangle) cut from the SAR dataset shown in Fig. 13. SAR image #2 is $1455 \times 1707$ in size, and 6 ships are visible. The sea waves make the SAR background look rough.

The statistical analysis of SAR image #2 is shown in Fig. 15. Similarly, the pixels of the 6 ship bodies in Fig. 14 are discarded from the SAR image before statistical distribution fitting is carried out. A histogram of the pixels of the clutter background in SAR image #2 is given in Fig. 15, and the PDF curves of the Gaussian, Weibull, Rayleigh, K and Gamma distributions used to fit the background clutter data are also included. Additionally, the tails of these statistical distributions of SAR image #2 are given in Fig. 16. In contrast to those of SAR image #1, both the main body and tail of the PDF of the K distribution fit the histogram of SAR image #2 very well, whereas the Gaussian distribution greatly deviates from the histogram of SAR image #2, as expected. The tail of the Rayleigh distribution is lower than that of the histogram of SAR image #2, and the tail of the Weibull distribution is slightly lower than that of the histogram. The tails of the PDFs of both Gamma distributions are higher than that of the histogram of SAR image #2 to some degree. The goodness of fit of these statistical distributions to SAR image #2 is different from that to SAR image #1.

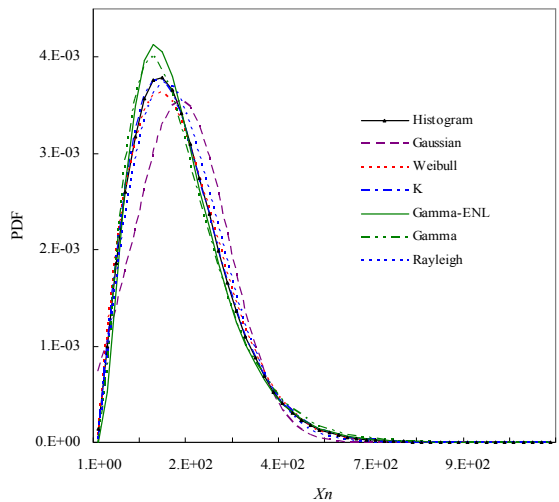

**Fig. 15.** The analysis of the statistical distribution of SAR image #2.

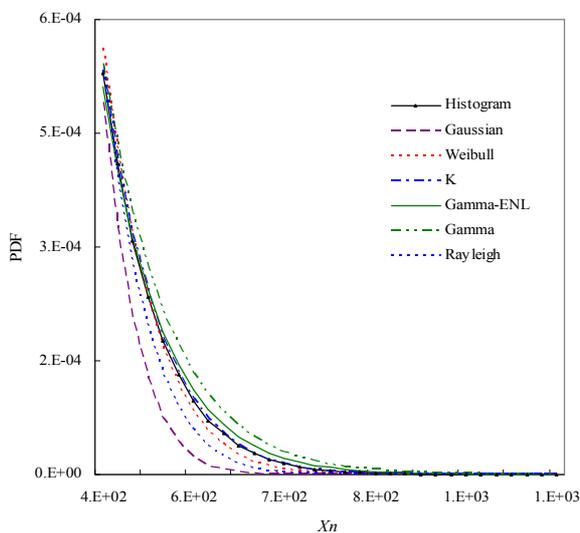

**Fig. 16.** The tails of the statistical distribution of SAR image #2.

### 2) **Performance analysis of the Wilcoxon nonparametric detector**

Here, the Wilcoxon nonparametric detector takes a test window of size $t \times t = 2 \times 2$, three boundary rows/columns $q=3$ of reference samples, a width $g=60$ of the guard area, and a total number of reference samples of 1500. The Wilcoxon nonparametric detector slides 2 pixels for each detection. The two-parameter CFAR, the Weibull-CFAR and the TS-CFAR take a single pixel in the test window, a single boundary row/column $q=1$ of reference samples, and the same width $g=60$ of the guard area. The total number of reference samples is 488. For the TS-CFAR scheme, the truncation ratio is $R_t = 10\%$. The AIS-RCFAR does not use the guard window, and it takes the same reference window of size $\ell \times \ell = 123 \times 123$ as the other parametric CFAR schemes. The parameter $\lambda$ of the AIS-RCFAR is set to $\lambda=2.0$.

The detection results of the two-parameter CFAR on SAR image #2 are shown in Figs. 17(a), (b) and (c) for the design false alarm rates $P_{FA}=10^{-6}$, $10^{-7}$ and $10^{-8}$, respectively. The detection results of the Weibull-CFAR on SAR image #2 are given in Figs. 18(a), (b) and (c) for the design false alarm rates $P_{FA}=10^{-3}$, $10^{-4}$ and $10^{-5}$, respectively. The detection results of the TS-CFAR on SAR image #2 are shown in Fig. 19(a), (b) and (c) for the design false alarm rates $P_{FA}=10^{-2}$, $10^{-3}$ and $3 \times 10^{-4}$, respectively. The detection results of the AIS-RCFAR on SAR image #2 are presented in Figs. 20(a), (b) and (c) for the design false alarm rates $P_{FA}=10^{-4}$, $10^{-5}$ and $10^{-6}$, respectively. The detection results of the Wilcoxon CFAR on SAR image #2 at the design false alarm rates $P_{FA}=10^{-6}$, $10^{-7}$ and $10^{-8}$ are given in Figs. 21(a), (b) and (c), respectively.

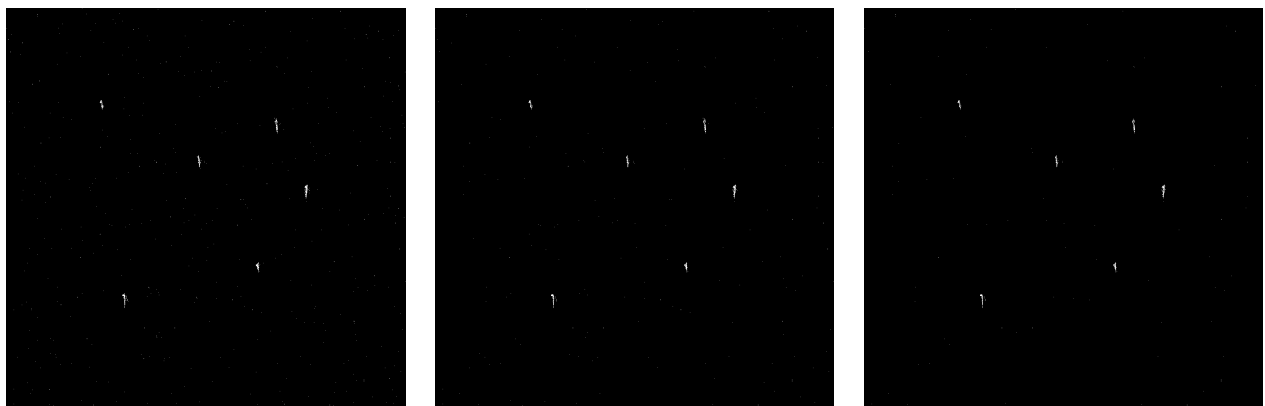

(a) $P_{FA}=10^{-6}$, $P_{fa}=3.67 \times 10^{-4}$      (b) $P_{FA}=10^{-7}$, $P_{fa}=1.7 \times 10^{-4}$      (c) $P_{FA}=10^{-8}$, $P_{fa}=0.90 \times 10^{-4}$

**Fig. 17.** The detection results of the two-parameter CFAR on SAR image #2.

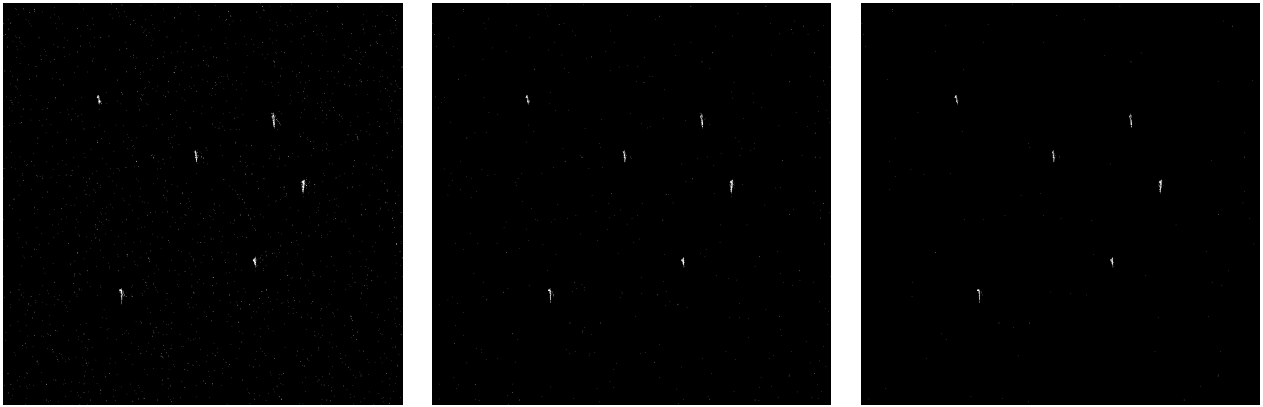

(a) $P_{FA}=10^{-3}$, $P_{fa}=1.4\times10^{-3}$        (b) $P_{FA}=10^{-4}$, $P_{fa}=3.35\times10^{-4}$        (c) $P_{FA}=10^{-5}$, $P_{fa}=0.88\times10^{-4}$

**Fig. 18.** The detection results of the Weibull-CFAR on SAR image #2.

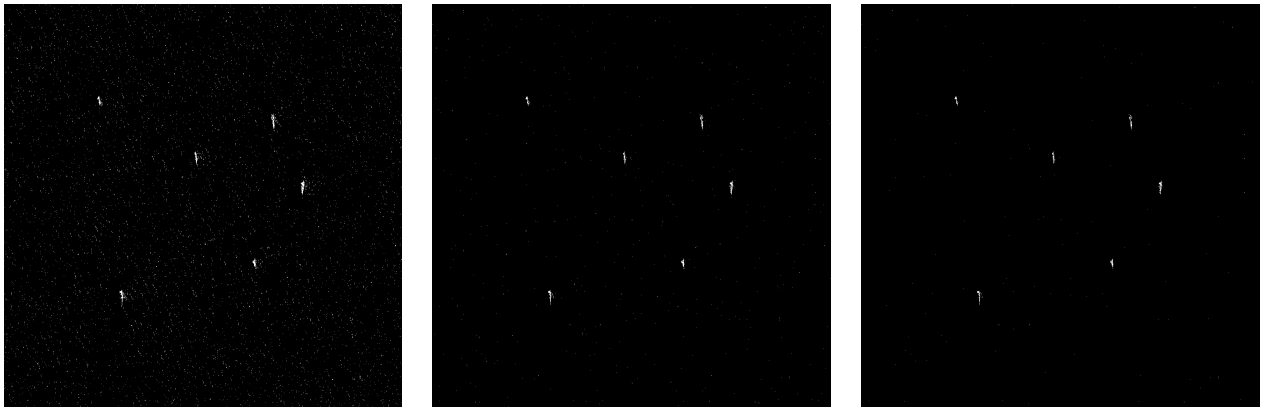

(a) $P_{FA}=10^{-2}$, $P_{fa}=4.4\times10^{-3}$        b) $P_{FA}=10^{-3}$, $P_{fa}=3.91\times10^{-4}$        (c) $P_{FA}=3\times10^{-4}$, $P_{fa}=1.07\times10^{-4}$

**Fig. 19.** The detection results of the TS-CFAR scheme on SAR image #2.

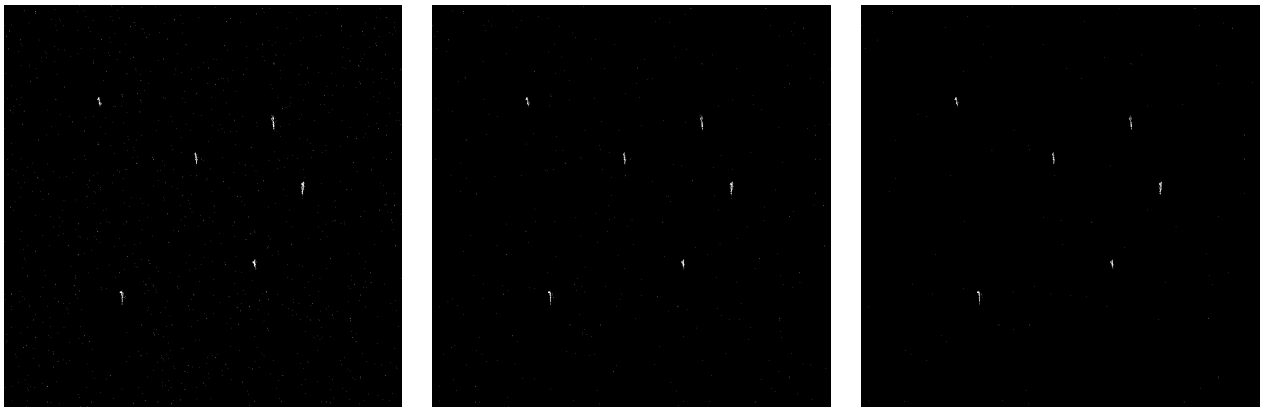

(a) $P_{FA}=10^{-4}$, $P_{fa}=8.25\times10^{-4}$        (b) $P_{FA}=10^{-5}$, $P_{fa}=2.63\times10^{-4}$        (c) $P_{FA}=10^{-6}$, $P_{fa}=0.86\times10^{-4}$

**Fig. 20.** The detection results of the AIS-RCFAR scheme on SAR image #2.

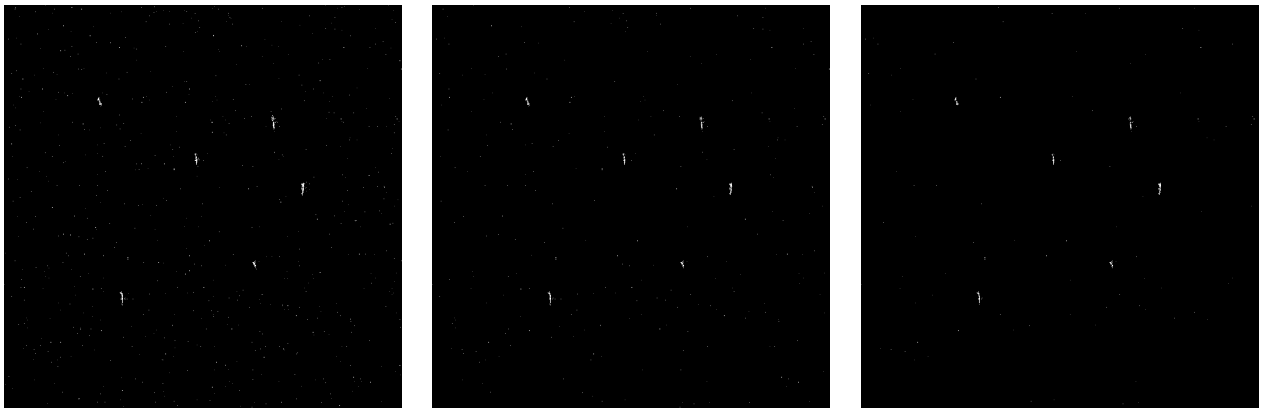

(a) $P_{FA}=10^{-6}$, $P_{fa}=2.72\times10^{-4}$  (b) $P_{FA}=10^{-7}$, $P_{fa}=1.13\times10^{-4}$  (c) $P_{FA}=10^{-8}$, $P_{fa}=0.44\times10^{-4}$

**Fig. 21.** The detection results of the Wilcoxon nonparametric detector on SAR image #2.

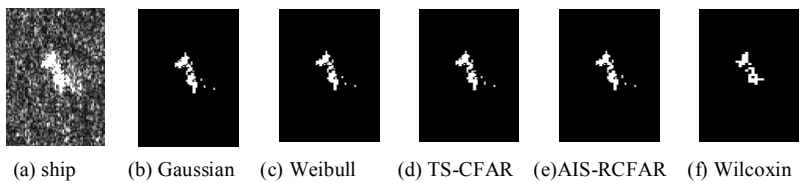

(a) ship  (b) Gaussian  (c) Weibull  (d) TS-CFAR  (e)AIS-RCFAR  (f) Wilcoxin

**Fig. 22.** The detection effects of several detectors on ship in SAR image #2.

It can be seen that for the detection results of these detectors on SAR image #2 in Fig. 17(c) through Fig. 21(c), although the design false alarm rates $P_{FA}$ of the two-parameter CFAR, the Weibull-CFAR, the TS-CFAR, the AIS-RCFAR and the Wilcoxon nonparametric detector are different, they achieve similar actual false alarm rates $P_{fa}$ of an acceptable level of $10^{-4}$. The values of $N_{fa}$, $N_c$, $P_{fa}$ and $P_{FA}$ for the detection results for these detectors in Fig. 17(c) through Fig. 21(c) are presented in Table I.

The detection effects on the ship in the white circle from SAR image #2 by the two-parameter CFAR in Fig. 17(c), the Weibull-CFAR in Fig. 18(c), the TS-CFAR in Fig. 19(c), the AIS-RCFAR in Fig. 20(c) and the Wilcoxon nonparametric CFAR in Fig. 21(c) are illustrated in Fig. 22 again. It can be seen that for large or medium ship targets in SAR images that have strong reflection effects, these detectors can detect most of the pixels in the ship body under a similar actual false alarm rate $P_{fa}$.

3) **The time cost of the Wilcoxon nonparametric detector**

The execution times $T_s$ of the two-parameter CFAR, the Weibull CFAR, the TS-CFAR, the AIS-RCFAR and the Wilcoxon nonparametric detector on SAR image #2 in Fig. 17 (c) through Fig. 21(c) are also given in Table I. Among these CFAR schemes, ship detection on SAR image #2 by the Wilcoxon nonparametric detector is the fastest, and it reduces the execution time of the two-parameter CFAR by nearly half. The execution time of the TS-CFAR is the longest. This result is similar to that for SAR image #1.

*C. Weak ship detection in a rough SAR background*

1) **SAR Image #3 collected by Gaofen-3**

The SAR image shown in Fig. 23 is a scene of the AIR-SARShip-1.0 dataset [27]. The AIR-SARShip-1.0 dataset was collected by the Gaofen-3 satellite, which is the first Chinese C-band multi-polarization high-resolution synthetic aperture radar satellite. The image has a resolution of 3 m in single polarization mode and was saved in the "tiff" format. Fig. 24 shows a slice of SAR image #3 cut from the SAR scene (marked with a white rectangle) shown in Fig. 23. The SAR image #3 is 1201×1101 in size. There are 2 obvious ships and a weak ship (in a red circle) in SAR image #3. The sea condition is Level 4, and the ripples caused by the waves are clearly visible.

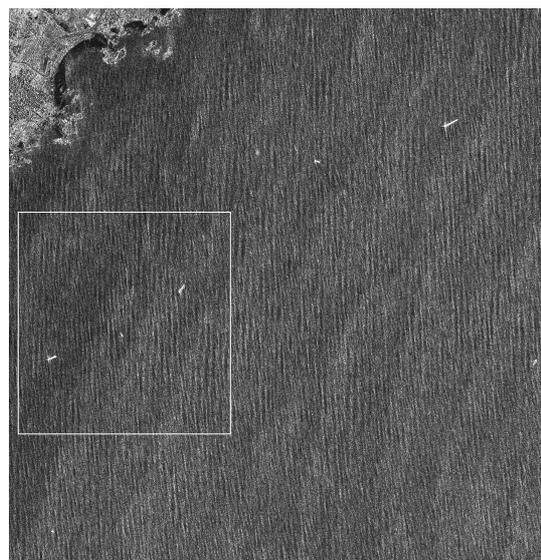

**Fig. 23.** The SAR image is a scene of the AIR-SARShip-1.0 dataset acquired by the C-band Gaofen-3.

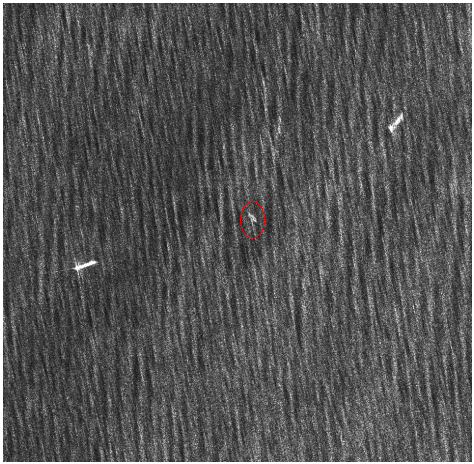

**Fig. 24.** SAR image #3 from Gaofen-3.

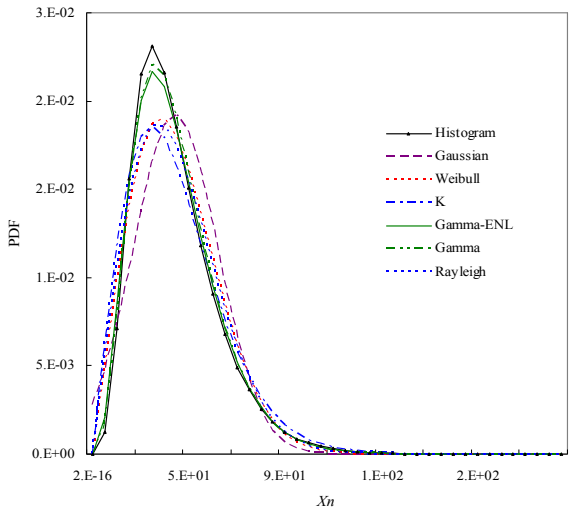

**Fig. 25.** The analysis of the statistical distribution of SAR image #3.

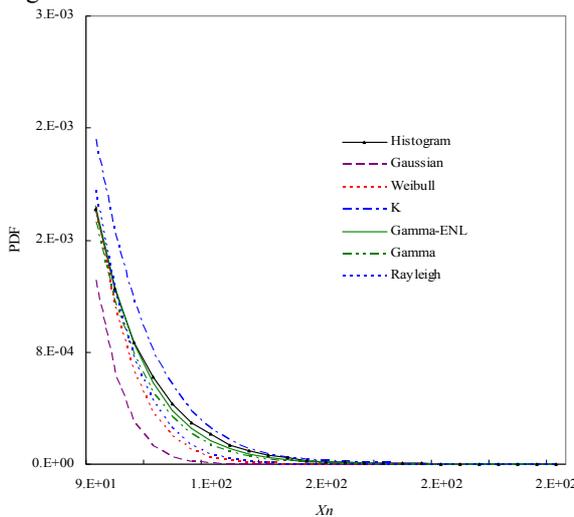

**Fig. 26.** The tails of the statistical distribution of SAR image #3.

To model the statistical distribution of SAR image #3, the ships in the SAR image are discarded as before. The statistical analysis of SAR image #3 is presented in Fig. 25. The tails of the statistical distributions of SAR image #3 are illustrated in Fig. 26. The statistical distribution fitting for SAR image #3 differs from that for the former two cases. None of these distribution models fit the histogram of the SAR image #3 very well. For the tails of the distribution, the closest to the tail of the histogram for SAR image #3 is that of the Gamma distribution with ENL, that of Gamma by the maximum likelihood method is slightly lower than it, and that of the Rayleigh and that of the Weibull are even lower. The tail of the Gaussian deviates greatly from the tail of the histogram for SAR image #3, and the tail of the K distribution is slightly higher than that of the histogram. Therefore, a specific statistical distribution may fit a certain SAR scene well, but if the SAR detection scene turns a different one, the assumed distribution may deviate from the actual SAR background such that the false alarm performance of a parametric CFAR scheme may deteriorate. The false alarm rate of the Wilcoxon nonparametric CFAR is independent of the distribution type of the SAR image background. This is the main motivation for us to apply the Wilcoxon nonparametric CFAR for ship detection in SAR images.

### 3) Performance analysis of the Wilcoxon nonparametric detector

Here, we verify the detection effects of the Wilcoxon nonparametric CFAR for weak ship against rough sea backgrounds, as shown in SAR image #3. The two-parameter CFAR, the Weibull-CFAR and the TS-CFAR take a single pixel in the test window, a single boundary row/column $q$=1 of reference samples, and a width $g$=60 of the guard area. The total number of reference samples is 488 for them. For the TS-CFAR scheme, the truncation ratio is also $R_t$ = 10% . The Wilcoxon nonparametric detector takes a test window of size $t \times t$=2 × 2, three boundary rows/columns $q$=3 of reference samples, a width $g$=60 of the guard area, and a total number of reference samples of 1500. The AIS-RCFAR does not use the guard window, and it takes a reference window of size $\ell \times \ell$ = 123×123 as the other parametric CFAR schemes. The parameter λ of the AIS-RCFAR is set to λ=2.0. To detect more details of the weak ship body, the Wilcoxon nonparametric CFAR slides a pixel for each detection here.

The detection results of the two-parameter CFAR on SAR image #3 are shown in Figs. 27(a), (b) and (c) for the design false alarm rates $P_{FA}$=10$^{-9}$, 10$^{-10}$ and 10$^{-11}$, respectively. The detection results of the Weibull-CFAR on SAR image #3 are shown in Figs. 28(a), (b) and (c) for the design false alarm rates $P_{FA}$=10$^{-6}$, 10$^{-7}$ and 3×10$^{-8}$, respectively. The detection results of the TS-CFAR on SAR image #3 are shown in Figs. 29(a), (b) and (c) for the design false alarm rates $P_{FA}$=10$^{-3}$, 10$^{-4}$ and 3×10$^{-5}$, respectively. The detection results of the AIS-RCFAR on SAR image #3 are shown in Figs. 30(a), (b) and (c) for the design false alarm rates $P_{FA}$=10$^{-5}$, 10$^{-6}$ and 4×10$^{-7}$, respectively. The detection results of the Wilcoxon detector on SAR image #3 at the design false alarm rates $P_{FA}$=10$^{-6}$, 10$^{-7}$ and $10^{-8}$ are shown in Figs. 31(a), (b) and (c), respectively. It can be observed from the detection results of these detectors on SAR image #3 in Fig. 27(c) through Fig. 31(c) that although the design false alarm rates $P_{FA}$ of the two-parameter CFAR, the Weibull CFAR, the TS-CFAR, the AIS-RCFAR and the Wilcoxon nonparametric detector are different, they reach a similar actual false alarm rate $P_{fa}$ of $10^{-4}$ level. The values of $N_{fai}$, $N_c$, $P_{fai}$ and $P_{FA}$ for the detection results by these detectors in Fig. 27(c) through Fig. 31(c) are presented in Table I.

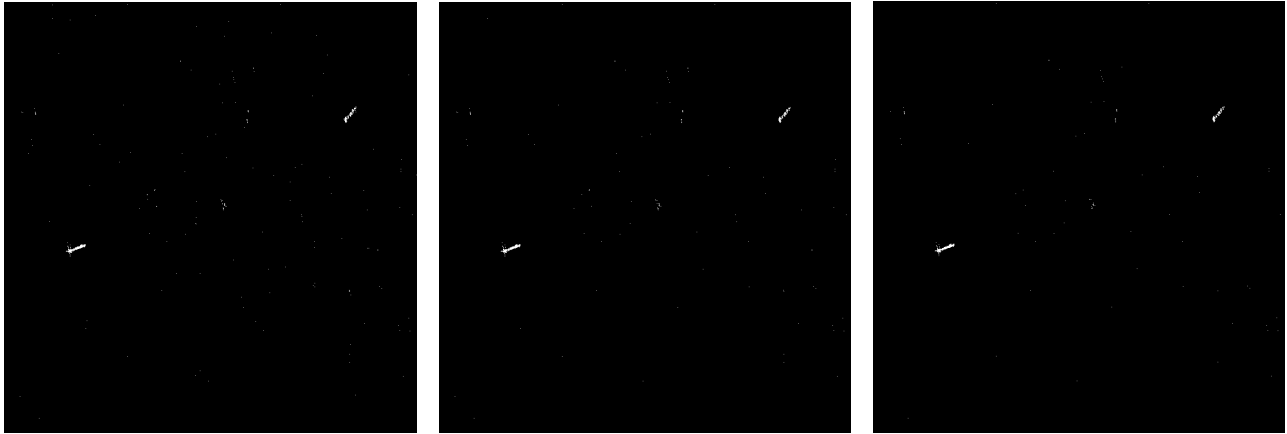

(a) $P_{FA}=10^{-9}$, $P_{fa}=2.32\times10^{-4}$      (b) $P_{FA}=10^{-10}$, $P_{fa}=1.4\times10^{-4}$      (c) $P_{FA}=10^{-11}$, $P_{fa}=1.04\times10^{-4}$

**Fig. 27.** The detection results of the two-parameter CFAR on SAR image #3.

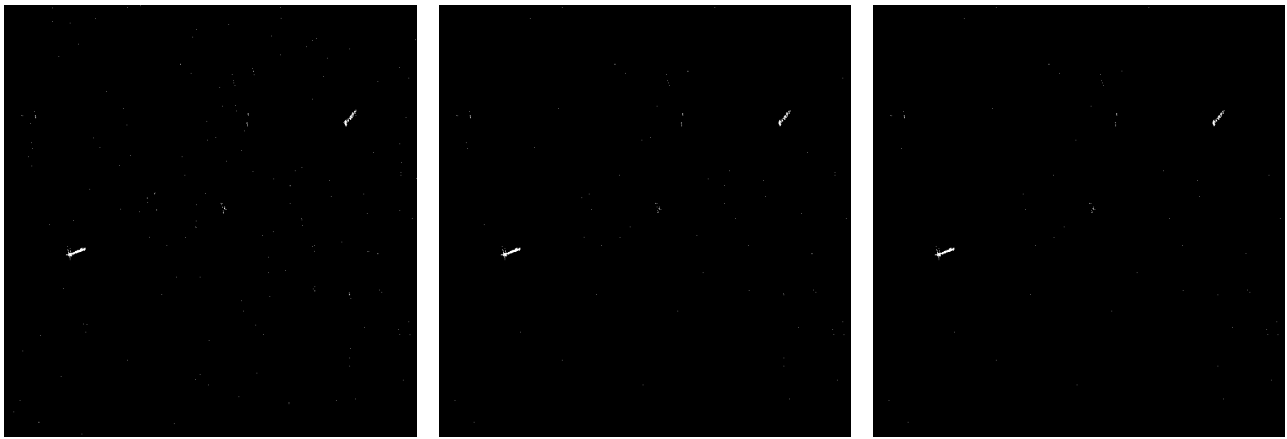

(a) $P_{FA}=10^{-6}$, $P_{fa}=2.86\times10^{-4}$      (b) $P_{FA}=10^{-7}$, $P_{fa}=1.33\times10^{-4}$      (c) $P_{FA}=3\times10^{-8}$, $P_{fa}=1.03\times10^{-4}$

**Fig. 28.** The detection results of the Weibull-CFAR on SAR image #3.

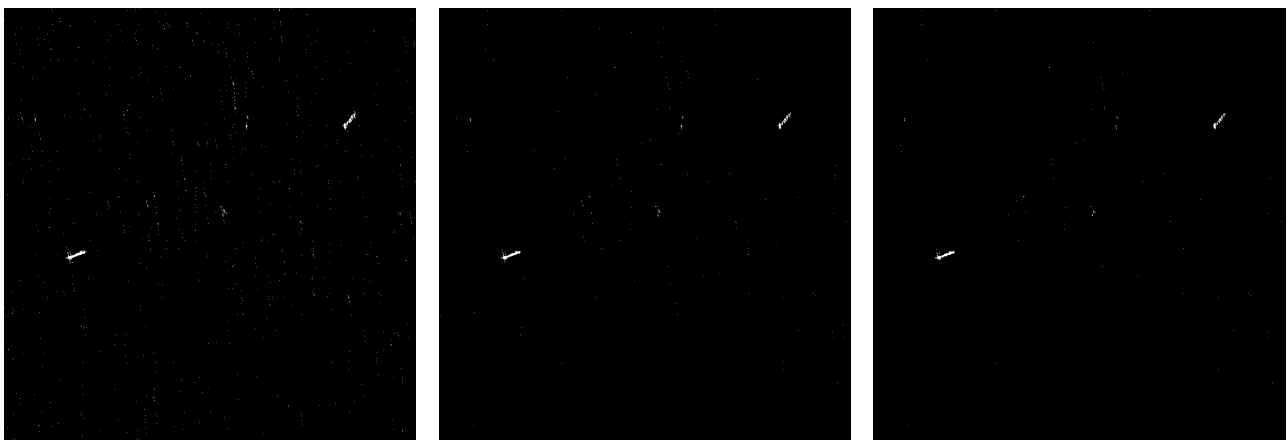

(a) $P_{FA}=10^{-3}$, $P_{fa}=1.15\times10^{-3}$      (b) $P_{FA}=10^{-4}$, $P_{fa}=2.09\times10^{-4}$      (c) $P_{FA}=3\times10^{-5}$, $P_{fa}=1.01\times10^{-4}$

**Fig. 29.** The detection results of the TS-CFAR on SAR image #3.

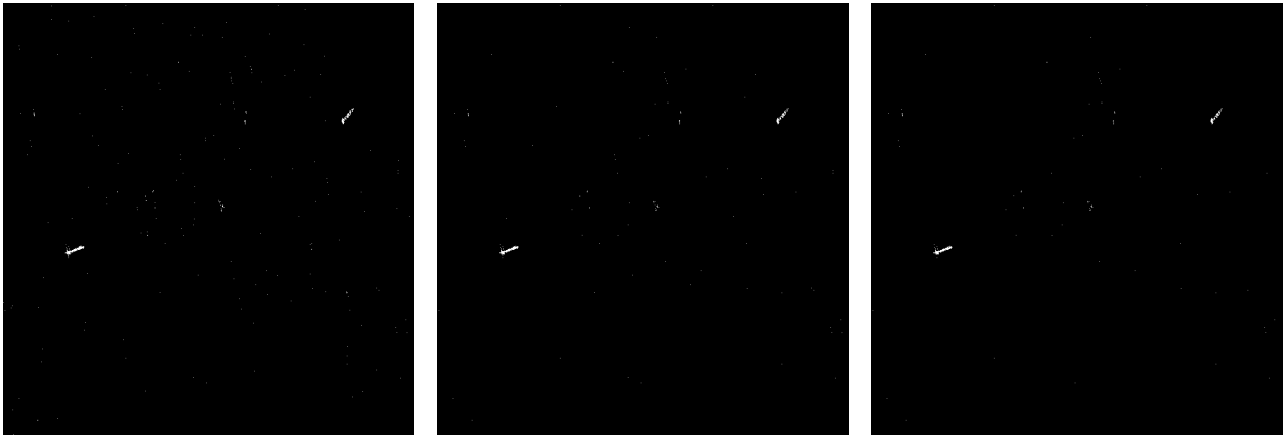

(a) $P_{FA}=10^{-5}$, $P_{fa}=2.98\times10^{-4}$       (b) $P_{FA}=10^{-6}$, $P_{fa}=1.34\times10^{-4}$       (c) $P_{FA}=4\times10^{-7}$, $P_{fa}=1.03\times10^{-4}$

**Fig. 30.** The detection results of the AIS-RCFAR on SAR image #3.

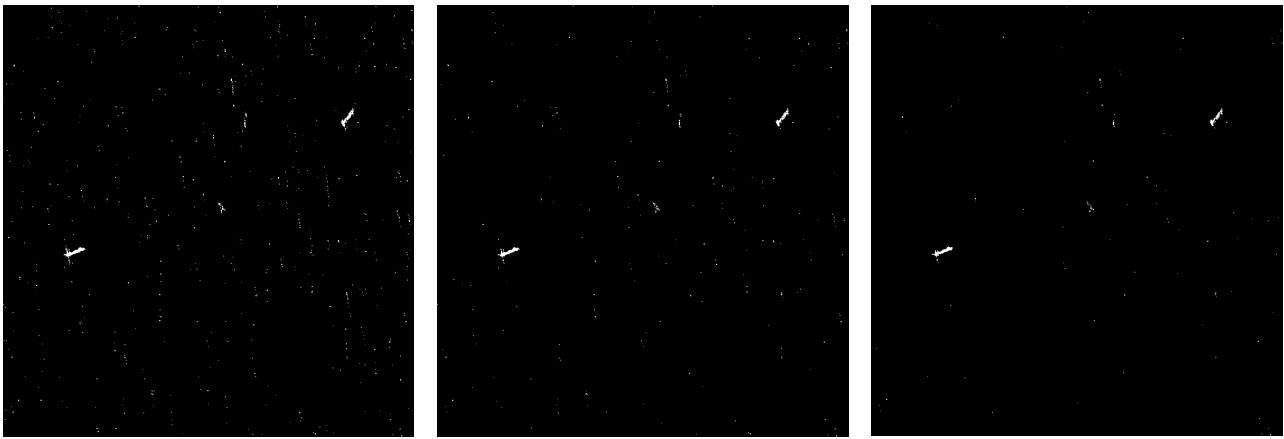

(a) $P_{FA}=10^{-6}$, $P_{fa}=4.41\times10^{-4}$       (b) $P_{FA}=10^{-7}$, $P_{fa}=1.64\times10^{-4}$       (c) $P_{FA}=10^{-8}$, $P_{fa}=0.68\times10^{-4}$

**Fig. 31.** The detection results of the Wilcoxon nonparametric detector on SAR image #3.

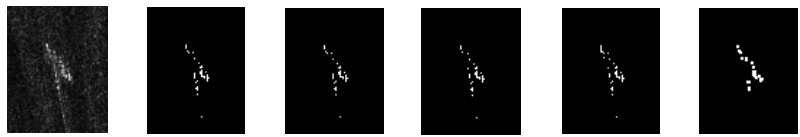

(a) ship    (b) Gaussian    (c) Weibull    (d) TS-CFAR    (e)AIS-RCFAR    (f) Wilcoxin

**Fig. 32.** The detection effects of several detectors on weak ship in SAR image #3.

To clearly observe the detection effects of the Wilcoxon nonparametric CFAR on the weak ship in the red circle in SAR image #3, the detection results of these detectors on the weak ship in Fig. 27(c) through Fig. 31(c) are illustrated in Fig. 32 again. More pixels in the ship body are detected by the Wilcoxon nonparametric detector than by the two-parameter detector, the Weibull-CFAR, the TS-CFAR and the AIS-RCFAR.

### 3) **The time cost of the Wilcoxon nonparametric detector**

The execution times $T_s$ of these detectors on SAR image #3 in Fig. 27(c) through Fig. 31(c) are given in Table I. Among these CFAR schemes, the time cost of the two-parameter CFAR on SAR image #3 is the least, and that of the Wilcoxon nonparametric CFAR is approximately 2 times that of the two-parameter CFAR. This is because the Wilcoxon nonparametric CFAR slides one pixel for each detection here. The time cost of the TS-CFAR is the longest.

## IV. DISCUSSION

The first step in the performance analysis of a CFAR scheme is to evaluate its ability to control the false alarm rate to a suitably low level against different backgrounds. The values of $N_{fa}$, $N_c$, $P_{fa}$ and $P_{FA}$ for the detection results by the two-parameter CFAR, the Weibull-CFAR, the TS-CFAR, the AIS-RCFAR and the Wilcoxon nonparametric detector on SAR images #1, #2 and #3 in Fig. 6(c) through Fig. 10(c), Fig. 17(c) through Fig. 21(c), and Fig. 27(c) through Fig. 31(c) are summarized in Table I.

It is shown in Table I that to reach a similar actual false alarm rate $P_{fa}$ of $10^{-4}$ level, the design false alarm rates $P_{FA}$ of the two-parameter CFAR on SAR images #1, #2 and #3 change by 4 orders of magnitude, whereas those of the Weibull-CFAR vary by 3 orders of magnitude. In other words, if the design false alarm rates of the two-parameter CFAR and the Weibull-CFAR hold the same, the actual false alarm performance deteriorates greatly. The design false alarm rates $P_{FA}$ of the TS-CFAR and the AIS-RCFAR on SAR images #1,

#2 and #3 change by approximately 1 order of magnitude to obtain an actual false alarm rate $P_{fa}$ near the $10^{-4}$ level. In particular, the design false alarm rate $P_{FA}$ and the actual false alarm rate $P_{fa}$ for the TS-CFAR are close. This means that the Gamma distribution assumed by the TS-CFAR fits the sea backgrounds in different SAR images relatively well. This point of view was also pointed out by Ward [28].

TABLE I
The values of $N_{fa}$, $N_c$, $P_{fa}$ and $P_{FA}$ for the detection results by these CFAR detectors on SAR images #1, #2 and #3.

|  |  | TP-CFAR | Weibull CFAR | TS-CFAR | AIS-RCFAR | Wilcoxon |
|---|---|---|---|---|---|---|
| SAR image #1 | $N_{fa}$ | 102 | 106 | 122 | 118 | 26 |
|  | $N_c$ | 900318 | 900318 | 900318 | 900318 | 900318 |
|  | $P_{fa}$ | $1.13\times10^{-4}$ | $1.18\times10^{-4}$ | $1.36\times10^{-4}$ | $1.31\times10^{-4}$ | $0.29\times10^{-4}$ |
|  | $P_{FA}$ | $1.0\times10^{-7}$ | $3.0\times10^{-5}$ | $1.0\times10^{-4}$ | $3.0\times10^{-6}$ | $1.0\times10^{-8}$ |
|  | $T_s$(s) | 30.86 | 311.17 | 382.35 | 66.50 | 12.83 |
| SAR image #2 | $N_{fa}$ | 223 | 219 | 266 | 213 | 108 |
|  | $N_c$ | 2481960 | 2481960 | 2481960 | 2481960 | 2481960 |
|  | $P_{fa}$ | $0.90\times10^{-4}$ | $0.88\times10^{-4}$ | $1.07\times10^{-4}$ | $0.86\times10^{-4}$ | $0.44\times10^{-4}$ |
|  | $P_{FA}$ | $1.0\times10^{-8}$ | $1.0\times10^{-5}$ | $3.0\times10^{-4}$ | $1.0\times10^{-6}$ | $1.0\times10^{-8}$ |
|  | $T_s$(s) | 86.34 | 944.72 | 1196.44 | 546.77 | 56.54 |
| SAR image #3 | $N_{fa}$ | 137 | 136 | 134 | 136 | 90 |
|  | $N_c$ | 1320909 | 1320909 | 1320909 | 1320909 | 1320909 |
|  | $P_{fa}$ | $1.04\times10^{-4}$ | $1.03\times10^{-4}$ | $1.01\times10^{-4}$ | $1.03\times10^{-4}$ | $0.68\times10^{-4}$ |
|  | $P_{FA}$ | $1.0\times10^{-11}$ | $3.0\times10^{-8}$ | $3.0\times10^{-5}$ | $4.0\times10^{-7}$ | $1.0\times10^{-8}$ |
|  | $T_s$(s) | 47.30 | 485.04 | 644.80 | 393.23 | 118.68 |

It can also be observed that if the design false alarm rate of the Wilcoxon nonparametric CFAR is kept the same of $P_{FA}=10^{-8}$ across SAR images #1, #2 and #3, the actual false alarm rate $P_{fa}$ of the Wilcoxon nonparametric CFAR is confined below a level of $10^{-4}$. Thus, the Wilcoxon nonparametric detector exhibits a strong ability to control false alarm rates at a low level in these different SAR backgrounds. This is due to the fact that the false alarm rate of the Wilcoxon nonparametric is independent of the background distribution.

The second step in the performance analysis of a CFAR scheme is to analyze and compare its detection probability under the same or similar actual false alarm rates. For large or medium ship targets, which have strong reflection effects in the SAR images, all of these detectors can detect most of the pixels in the ship body under a similar actual false alarm rate $P_{fa}$. The advantage of detection performance for the Wilcoxon nonparametric CFAR cannot manifest in this case. However, the difference in the detection performance of these detectors for weak ship targets can be clearly identified. To quantitatively analyze the detection performance for a weak ship target, the detection probability $P_d$ is defined as

$$P_d = \frac{N_d}{N_s} \qquad (12)$$

where $N_d$ is the number of detected pixels in the ship body and $N_s$ is the total number of pixels in the ship body. We also use the ellipse to fit the ship body to estimate $N_s$ here.

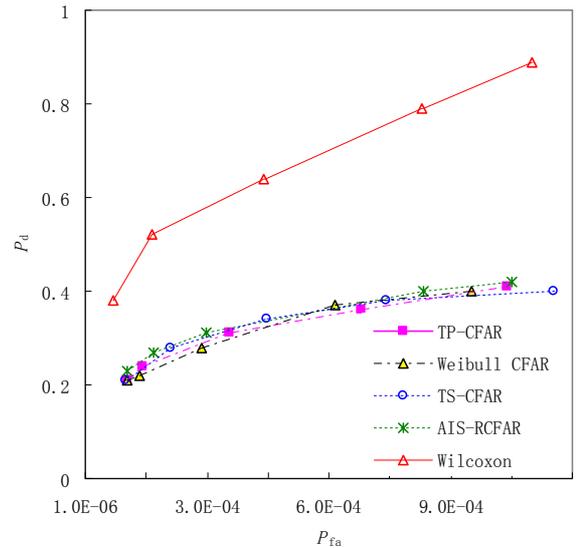

**Fig. 33.** The ROC curves of these several CFAR schemes.

For a given value of the design false alarm rate $P_{FA}$, the detection threshold of the corresponding CFAR scheme is determined first, and then target detection is performed on SAR image #3. We can calculate the actual false alarm rate $P_{fa}$ according to (11) and the detection probability $P_d$ for the weak ship in the red circle of SAR image #3 by (12). As such, several value pairs of the detection probability $P_d$ on the weak ship versus the actual false alarm rate $P_{fa}$ on SAR image #3 can be obtained. The receiver operating characteristic (ROC) curves of these CFAR schemes are illustrated in Fig. 33. Under a similar actual false alarm rate $P_{fa}$, the detection performances of these parametric CFAR schemes on the weak ship are close, and that of the Wilcoxon nonparametric CFAR exhibits an evident advantage over these parametric CFAR detectors.

## V. Conclusion

The ship detection in SAR images has become a hot topic in the field of marine remote sensing, and many papers focused on this topic have been published. Most of them are the parametric CFAR detection schemes that assume a known clutter distribution for the background of the SAR image. The parametric CFAR detector suffers from the disadvantage that the detection threshold is sensitive to changes in the background clutter distribution. The nonparametric CFAR detector has an inherent advantage in that it can maintain a constant false alarm rate even if the clutter distribution becomes different. The Wilcoxon nonparametric CFAR scheme for ship detection in SAR images is proposed and analyzed in this work. A closed form of the false alarm rate for the Wilcoxon nonparametric CFAR detector to determine the decision threshold is presented. A comparison between the detection performance of the Wilcoxon nonparametric detector, that of the two-parameter CFAR, that of the Weibull-CFAR, that of the AIS-RCFAR and that of the TS-CFAR is made, and our experimental tests are carried out on Radarsat-2, ICEYE-X6 and Gaofen-3 SAR images.

If the design false alarm rate of the Wilcoxon nonparametric CFAR is kept at the same $P_{FA} = 10^{-8}$ across SAR images #1, #2 and #3, the actual false alarm rate $P_{fa}$ in the detection results can be confined to below a level of $10^{-4}$. This means that the Wilcoxon nonparametric detector exhibits a strong ability to control the false alarm rates at a very low level against these different SAR backgrounds. This is due to the fact that the false alarm rate of the Wilcoxon nonparametric CFAR is independent of the background distribution.

Regarding the detection performance of the Wilcoxon nonparametric CFAR, under the similar actual false alarm rates $P_{fa}$ of $10^{-4}$ level, all of these detectors can detect most of the pixels in the ship body that have strong reflection effects. The advantage of detection performance for the Wilcoxon nonparametric CFAR is not revealed in this case. However, its advantage over these parametric CFAR schemes is evident for the detection of weak ship in SAR images. It should be noted that although the detection performances of these parametric CFAR schemes on weak ship under similar actual false alarm rates are close, they have significantly different capabilities for controlling false alarms against different SAR backgrounds.

Moreover, the Wilcoxon nonparametric detector can suppress the false alarms resulting from sidelobes to some extent. Additionally, the detection speed of the Wilcoxon nonparametric detector is fast, and it saves nearly half of the execution time of the two-parameter CFAR. The Wilcoxon nonparametric detector needs only $m \times n + 1$ comparators and an accumulator to be implemented; thus, it has simple hardware implementation. Since the Wilcoxon nonparametric detector does not assume a known distribution for the background clutter, it avoids the complexity and the large computation time of the maximum likelihood estimation on the distribution parameters. Therefore, the Wilcoxon nonparametric CFAR detector can be a good choice for ship detection in SAR images.

To enhance the robustness of the detection performance of the Wilcoxon nonparametric CFAR detector in nonhomogeneous backgrounds caused by multiple targets and clutter boundaries, the Wilcoxon nonparametric CFAR detector will be integrated with adaptive censoring techniques, such as the ordered data variable (ODV) [29]; with the sub-window selection methods, such as the greatest of (GO), the smallest of (SO) and the variable index (VI) CFAR [30]; and with the superpixel-level technique [31], [32] to propose some modified Wilcoxon nonparametric CFAR detectors for nonhomogeneous backgrounds. Moreover, we shall combine the Wilcoxon nonparametric CFAR detector with the RCNN and YOLO deep learning networks or their diverse versions to improve the radar target detection performance. We shall accomplish these tasks in the near future.

### Acknowledgment

I would like to express my thanks to the ICEYE sample dataset [26], High-resolution SAR Ship Detection Dataset: AIR-SARShip-1.0 [27] and Prof. Hongwei Liu with Xidian University for help with our experimental SAR images.

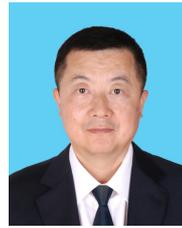

**Xiangwei Meng** was born in 1966. He graduated from Dalian University of Technology in 1987. He currently serves as a professor at Yantai Nanshan University. He is a member of IET and a senior member of CIE. He acts as a reviewer for IEEE Trans. on AES, IET RSN, Signal Processing and Electronics Letters. His research interests include radar CFAR detection and signal theory.